\documentclass[letterpaper, 10 pt, conference]{ieeeconf}  

\IEEEoverridecommandlockouts
\usepackage[vlined, ruled, boxed, linesnumbered]{algorithm2e}

\SetCommentSty{mycommfont}

\SetKwProg{Function}{}{}{}

\usepackage{graphicx,subfig}
\usepackage{graphics}
\usepackage{amssymb,amsmath}
\usepackage{amssymb}
\usepackage[noend]{algpseudocode}
\usepackage[font={small}]{caption}

\usepackage[usenames, dvipsnames]{color}

\usepackage[normalem]{ulem}

\usepackage{tikz}
\usepackage{textcomp}
\usepackage{lipsum}
\usepackage[noadjust]{cite}
\usepackage{gensymb}

\usepackage{booktabs}

\newcommand{\Ical}{\mathcal{I}}

\title{\LARGE \bf
DRACo-SLAM: Distributed Robust Acoustic Communication-efficient SLAM for Imaging Sonar Equipped Underwater Robot Teams
}

\vspace{-4mm}

\author{John McConnell, Yewei Huang, Paul Szenher, Ivana Collado-Gonzalez and Brendan Englot
\thanks{J. McConnell, Y. Huang, P. Szenher, I. Collado-Gonzalez and B. Englot are with Stevens Institute of Technology,
        Hoboken, NJ, 07030, USA
{\tt\scriptsize$\{$jmcconn1,yhuang85,pszenher,icollado,benglot$\}$@stevens.edu}}%
\thanks{This research was supported in part by ONR Grant N00014-21-1-2161 and NSF Grants IIS-1652064 and IIS-1723996.}
}

\begin{document}

\maketitle
\thispagestyle{empty}
\pagestyle{empty}

\begin{abstract}
An essential task for a multi-robot system is generating a common understanding of the environment and relative poses between robots. Cooperative tasks can be executed only when a vehicle has knowledge of its own state and the states of the team members. However, this has primarily been achieved with direct rendezvous between underwater robots, via  inter-robot ranging. 
We propose a novel distributed multi-robot simultaneous localization and mapping (SLAM) framework for underwater robots using imaging sonar-based perception. 
By passing only scene descriptors between robots, we do not need to pass raw sensor data unless there is a likelihood of inter-robot loop closure. We utilize pairwise consistent measurement set maximization (PCM), 
making our system robust to erroneous loop closures. The functionality of our system is demonstrated using two real-world datasets, one with three robots and another with two robots. We show that our system effectively estimates the trajectories of the multi-robot system and keeps the bandwidth requirements of inter-robot communication low. 
To our knowledge, this paper describes the first instance of multi-robot SLAM using real imaging sonar data (which we implement offline, using simulated communication). Code link: https://github.com/jake3991/DRACo-SLAM.
\end{abstract}

\vspace{-1.5mm}

\section{Introduction}
Underwater robotics has proliferated over the past decade, supporting a variety of tasks, including ship hull inspection, harbor security, maintenance-inspection-repair (MIR), and intelligence surveillance reconnaissance (ISR). However, autonomous underwater vehicles (AUVs) are limited in their sensing capabilities. Firstly, the global positioning system (GPS) cannot be used due to the attenuation of GPS signals in the water. While AUVs can surface to collect GPS measurements, this is inefficient and impossible during under-ice missions or certain tactical situations. Further, if a small, lightweight, low-power perceptual package is desired, underwater lidar may be infeasible, and the use of sonar may be required, especially in turbid or dark environments. While the capabilities of sonar-based systems have expanded to include fully autonomous operation, the primary area of contribution is still focused on single agents executing their missions. However, real-world problems often benefit from a team of robots, a multi-robot system. When considering tasks ranging from autonomous exploration to ISR, a multi-robot system can reduce time, energy expenditure or create redundancy when operating in an adversarial environment.

\begin{figure}[t]
\centering
\subfloat[Our experimental ``AUV": a custom-instrumented BlueROV2.
\label{fig:leading_rov}]{\includegraphics[height=3.8cm]{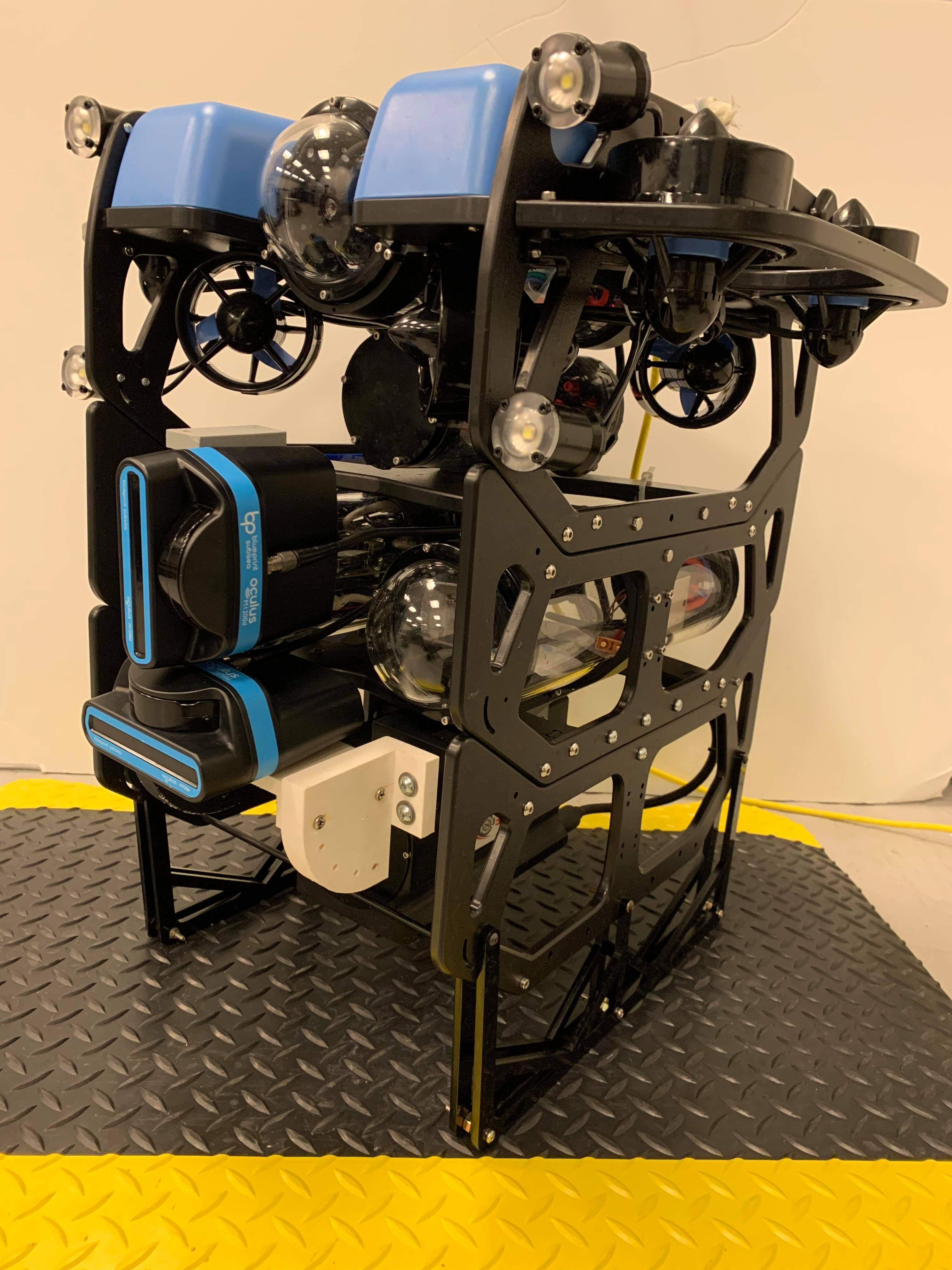}}\ \;
\subfloat[Satellite image of the SUNY Maritime marina.
\label{fig:leading_sat}]{\includegraphics[height=3.8cm]{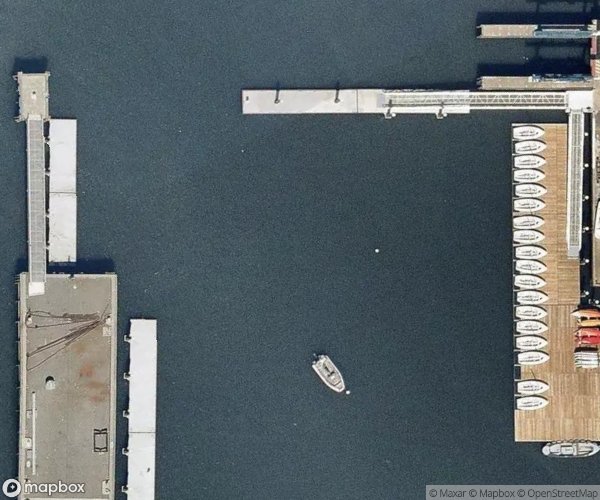}} \vspace{-5mm}\\
\subfloat[An example multi-robot SLAM run from the environment in (b), from the perspective of one robot. This robot's poses are shown as black dots connected by several factors: sequential scan matching factors (SSM) in green, non-sequential scan matching factors (NSSM, intra-robot loop closures) shown in red, inter-robot (IR) loop closures in blue and partner robot (PR) factors in purple. This robot's map is shown in red with blue points merged from the other robot in the mission. 
\label{fig:leading_1}]{\includegraphics[width=.95\linewidth]{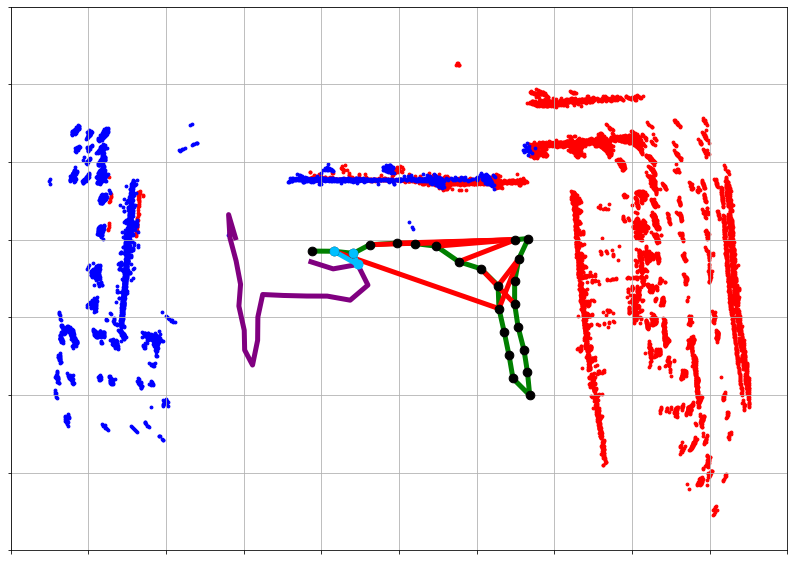}}\ \;
\caption{\textbf{Distributed multi-robot SLAM overview:} (a) the robot used in our experiments, (b) one of the settings for our experiments, and (c) a representative experimental result are shown.}
\vspace{-6mm}
\end{figure}

A fundamental capability for both single and multi-robot systems is state estimation. When operating in unknown environments, simultaneous localization and mapping (SLAM) is utilized to estimate vehicle location and provide situational awareness. For underwater robots, SLAM has typically been implemented to localize a single vehicle. 
Recent works on SLAM for underwater multi-robot systems consider both direct encounters (e.g., inter-robot range measurements) and indirect encounters (e.g., commonly observed targets in the survey area) as a means for fusing robot state estimates \cite{paull-2014,paull-2015}. There are some fundamental challenges to implementing such systems. Firstly, acoustic communication links have notably low bandwidth, limiting information exchange between robots. Another issue is the nature of sonar as a tool for perception. Sonar has a low signal-to-noise ratio, low resolution, and often lacks 3D information. Moreover, there are no (or poor) initial conditions relating the robots in a multi-robot system without GPS.

In this work, we propose a system for distributed multi-robot SLAM that uses an imaging sonar's perceptual data as a basis for state estimation, and we validate it over real-world datasets. 
In addition to utilizing real perceptual data, we do not provide any initial conditions relating the robots a priori, and we do not rely upon inter-robot ranging, permitting an exclusive reliance on indirect encounters between robots. Lastly, we consider the bandwidth limitations of wireless acoustic communications and design our system for compatibility with realistic  constraints (we note that beyond bandwidth, there are additional limitations associated with real-world acoustic comms \cite{acomms}). Our key contributions are: 
\begin{itemize}
    \item The first underwater multi-robot SLAM system developed for use with real imaging sonar data.
    \item A high-performance pipeline for robustly identifying, registering, and rejecting outliers of indirect encounters using real sonar data with no initial conditions.
    \item A strategy to manage bandwidth usage, making real world application with acoustic modems a possibility.  
    \item A validation of our system on two real-world data-sets. 
    
\end{itemize}

In the following sections, we discuss related work, mathematically define our problem, present our system and results.


\vspace{-1.75mm}

\section{Related Work}
\subsection{Underwater Multi-Robot SLAM}

In the underwater space, inter-robot constraints generally come in two forms \cite{paull-2015}: \textit{direct} encounters where robots observe one another via acoustic ranging, and \textit{indirect} encounters, where robots observe the same targets in the environment and may derive inter-robot measurement constraints relating one another. We note that direct encounters require synchronized clocks, which may not be practical over long periods without GPS clock corrections due to clock drift. 

We first consider \cite{fischell-2019}, where a mobile base station is used to localize a team of low-cost vehicles lacking perceptual sensors using an acoustic beacon. A similar concept is considered in \cite{yao-2009}, except instead of a team of robots, a leader-follower arrangement is used. \cite{li-2019} considers inter-robot ranging without fixed acoustic beacons and performs a simulation study comparing the use of fixed beacons to a cooperative localization solution. \cite{Halsted} considers an algorithm for processing inter-robot acoustic pulses in a distributed manner. \cite{paull-2014} proposes a pose-graph-based method for cooperative localization of a team of robots using dead-reckoning, GPS, and inter-robot ranging. The outcome of this work is a system where a robot maintains an understanding of its state, and the team's state. Moreover, when GPS measurements occur at the surface, their effect is shared across robots. \cite{paull-2015} integrates perception into the above, and commonly observed features are shared and integrated into robot pose-graphs. \cite{paull-2015} assumes that robots share point-landmark observations in their survey area, but the framework is only tested in simulation. Further, due to GPS, each robot is effectively localized in a common frame, enabling the sharing of range-bearing measurements to commonly observed targets without a need to solve the complex data association problem. Recall, GPS may not always be available due to under-ice operations or tactical situations (e.g., jamming). 
\cite{mangelson-2018} considers the robot map merging problem using only similarity in feature space, in this case, ship hull curvature. While this work is able to merge trajectories and lower the data transmission requirements between robots, it requires a highly descriptive feature vector. It is only validated over a single dimension, whereas we consider a 3DOF system. 

There are some notable examples of underwater multi-robot SLAM using cameras \cite{elibol-PRL,pfingsthorn-2010,font-2020,Xanthidis-2021} and cameras with other navigation sensors \cite{campos-2016}. Cameras, however are not robust in all water conditions. While these works are informative, because we operate in turbid conditions with low visibility, we do not consider them further. 

Another area to consider is the concept of multi-session SLAM, where a robot is provided with a prior SLAM run and merges the map it builds with the prior run as the mission progresses. This is examined for ship hull inspection \cite{ozeg,do-2020}, bathymetric mapping \cite{galceran-2013}, ship-wreck reconstruction \cite{williams-113} and environmental monitoring \cite{bryson-2013}. However, multi-session does not account for communicating information between robots, as it is a single agent system with prior information. Additionally, it may be the case that the robot has some knowledge of its location in the prior map, constraining the inter-robot loop closure search space.  

In contrast to the above, 
we utilize no notion of an initial guess relating robot reference frames. Second, we consider the bandwidth limitations of real acoustic modem hardware and take steps to manage network utilization.
Lastly, we implement a fully functioning system to detect and estimate indirect encounters in sonar data: inter-robot loop closures. 

\vspace{-1.65mm}

\subsection{Place Recognition With Sonar}
Place recognition (loop closure) is fundamental for a perception-driven multi-robot system. Place recognition has been widely studied in LIDAR sensing \cite{context,intensity-context,context-plus}. These works assemble a 2D bird's eye view image of a 3D LIDAR scan with coarse discretization to compare scenes. They also derive a 1D descriptor to support scene search and retrieval. 

Place recognition has also been studied with sonar-equipped AUVs. Firstly, \cite{santos-2018} considers building scene graphs to compare scenes and evaluate rigid-body transformations but requires at least fifteen objects in each scene to run. Machine learning has also been used to support this task in \cite{ribeiro-2018,larsson-2020}. However, few public sonar datasets exist, and this is often a research area unto itself. Iterative closest point (ICP) based loop closure is used in our prior work \cite{Wang-2021} to support single-agent active SLAM. This work uses sonar derived point clouds with ICP; when ICP provides a transform between keyframes, we estimate overlap between the point clouds. Point cloud overlap, then pairwise consistent measurement set maximization (PCM) \cite{pcm}, are used to reject outlier loop closures. Inliers are integrated into the graph-based pose SLAM solution. We note the extension of the PCM algorithm  \cite{do-2020}, but due to its additional complexity, we will not utilize it in our work.  

\vspace{-1mm}

\subsection{Acoustic Modems}
To communicate, AUVs generally use acoustic modems, which are low-bandwidth compared to in-air systems. One of the most well-known devices is the WHOI Micromodem \cite{whoi-modem}, which can transfer a maximum of 5400 bits/s over a long-range. For shorter-range transmission (300 meters), higher bandwidth (62.5 kbits/s) units are available \cite{evo}. Moreover, recent research has demonstrated the feasibility of achieving even higher bandwidth in real-world conditions \cite{modem}. In our work, we consider bandwidth to be limited, and we will study the network utilization of our multi-robot system as a critical parameter. Further, we will take steps to minimize the transmission of large data structures.


\section{Problem Description} 
In this work we consider a team of robots $N$.  
Each robot $n \in N$ maintains its own SLAM solution, in its own reference frame $\Ical_n$. 
Each robot receives a set of observations $\mathbf z_{n,t}$  
at a given pose $\mathbf x_{n,t}$ 
across discrete time steps $t$. Each robot pose is defined in the plane (fixed depth) as
\begin{align}
    \mathbf x_{n,t}
    = \begin{pmatrix} x_r\ 
    y_r\
    \theta_r \end{pmatrix}^\top.
\end{align}
Each set of observations consists of sonar returns in spherical coordinates with ranges $r\in \mathbb{R}_+$, bearings $\theta \subseteq [-\pi,\pi)$, elevation angles $\phi \subseteq [-\pi,\pi)$, and associated intensity values $\gamma \in \mathbb{R}_+$. These sonar observations can be mapped into Cartesian coordinates by
\begin{align}
    \begin{pmatrix}  x \\  y \\  z \end{pmatrix} 
    = r\begin{pmatrix} \cos{\phi} \cos{\theta} \\ 
    \cos{\phi}\sin{\theta} \\ 
    \sin{\phi} \end{pmatrix}.
    \label{eq:to_cartesian}
\end{align}
Each robot moves through the environment according to the dynamics
\begin{align}
    \mathbf x_{n,t} = \mathbf g(\mathbf u_{n,t} , \mathbf x_{n,t-1}) + \sigma_{n,t},
\end{align}
where $\mathbf x_{n,t-1}$ is the previous timestep’s pose, $\mathbf u_{n,t}$ is the control command and $\sigma_{n,t}$ is process noise. The posterior probability over the time history of poses is defined as
\begin{align}
    p(\mathbf x_{n,1:t} , \mathbf m_n | \mathbf z_{n,1:t}, \mathbf u_{n,1:t}  ),
    \label{eq:post}
\end{align}
with map $\mathbf m_n$. While we consider a distributed multi-robot SLAM solution, there is no initial transform between each of the reference frames $\Ical_n$. 
This work aims to use each robot's observations to derive inter-robot loop closures and estimate team member trajectories in each robot's reference frame. 

\section{Algorithm} 
\begin{figure}[t]
\centering
{\includegraphics[width = .9\linewidth]{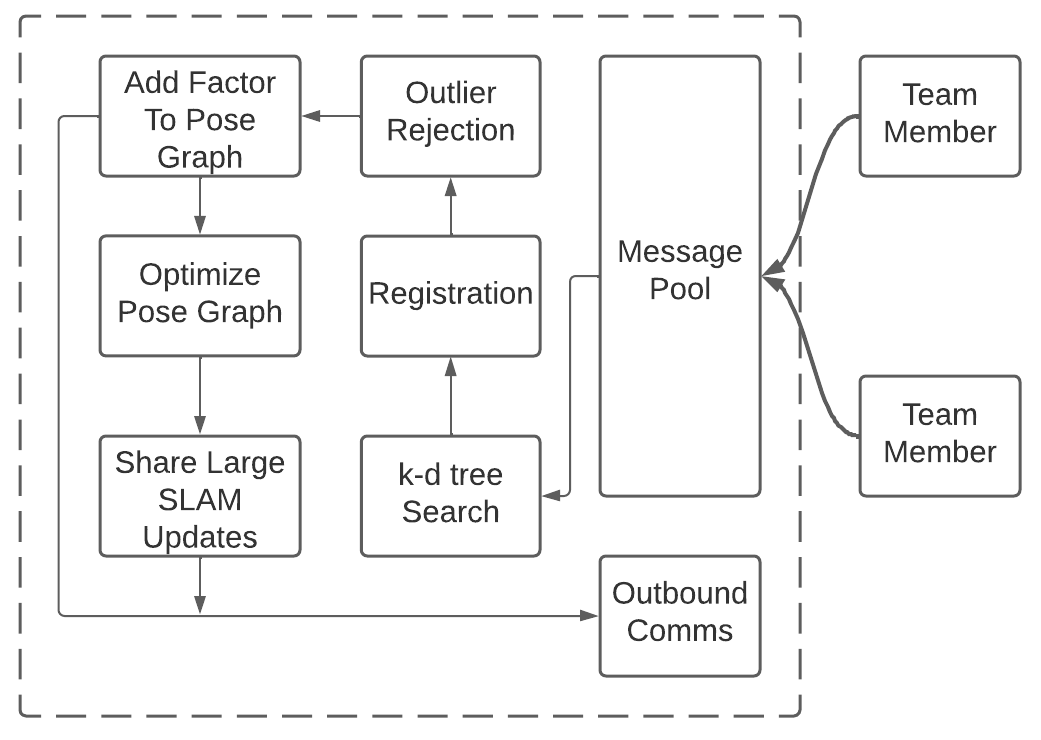}}
\caption{\textbf{System diagram for a single robot.} Each robot maintains a message pool of received scene descriptors. Based on k-d tree search, registration is attempted and outliers rejected. Inlier inter-robot loop closures are added to the pose graph and sent to team members. Large SLAM corrections are also sent to team members.  
}
\vspace{-6mm}
\label{fig:block_diagram}
\end{figure}

This section will provide the technical details of our distributed multi-robot SLAM system. An overview of this system from the perspective of one robot is shown in Fig. \ref{fig:block_diagram}. Each robot logs messages from team members, using a k-d tree to search for potential inter-robot loop closures. We then attempt registration and reject outliers. Inter-robot loop closures are added to the factor graph and broadcast to the rest of the team. The pose graph is optimized, and significant state changes are communicated to the rest of the team.

\subsection{Sonar image processing}
At each pose, $\mathbf x_{n,t}$, the sonar observations $\mathbf z_{n,t}$ consist of a sonar image. The 2D image is populated with acoustic intensity values. However, not all the pixels in the image represent contact with structures in the environment. Our first step is to identify which pixels constitute a sonar contact and which do not. We use constant false alarm rate (CFAR) detection, \cite{richards-2005} which is derived from radar processing and has supported our previous work \cite{ICRA-2021}. 


Once contacts are identified, the pixels are mapped to meters as an in-plane point cloud. Note that while imaging sonar observes a 3D volume of water, the sensor does not return 3D information, only a 2D projection with $\phi$ as zero. The consequence is that our system is confined to fixed depth, in-plane pose estimation. The point cloud is subject to voxel down-sampling, with each voxel's output being the medoid of the contained points. Example point clouds are shown in Fig. \ref{fig:leading_1}.

\subsection{Point Cloud Compression}
In this work, each point in a point cloud is a pair of 32-bit floating points (x,y), which may overwhelm the data link. For this reason, we consider a simple voxel-based compression algorithm. We take a point cloud, cast it onto a planar voxel grid, and only retain centers of occupied voxels. 
This compression is performed first by discretizing each point, $p_{x,y}$, to a designated resolution, $\Delta_{compression}$.
\begin{align}
    p_{i,j} \thickapprox \dfrac{p_{x,y}}{\Delta_{compression}}
    \label{eq:compression}
\end{align}
Note that the $p_{i,j}$ are discrete. Each cell in the voxel grid is populated according to Eq. (\ref{eq:compression}). If a voxel contains a non-zero number of points, then that voxel's center is recorded for transmission. In our implementation, each voxel center is a pair of 8-bit unsigned integers, offering significant savings over a pair of 32-bit floating points. This compression, though, results in a loss of some geometric information. Our experiments will evaluate compression efficacy by characterizing SLAM error and network utilization in Section V. 

\vspace{-1mm}

\subsection{Single robot SLAM}
For multi-robot SLAM, first, each robot must estimate its state. We utilize the vehicles' onboard Doppler velocity log (DVL) and inertial measurement unit (IMU) with the uncompressed point clouds from Section IV-A. We formulate this as a graph-based pose SLAM problem and use the GTSAM \cite{GTSAM} implementation of iSAM2 \cite{Kaess-2011}. We use odometry factors $\mathbf  f^{\text{0}}$ from the vehicle dead reckoning system between sequential poses. Next, we add sequential scan matching (SSM) factors. $\mathbf  f^{\text{SSM}}$ are derived from calling ICP \cite{icp} between sequential frames. Lastly, we consider non-sequential scan matching factors (i.e., intra-robot loop closures). $\mathbf  f^{\text{NSSM}}$  factors are derived by calling ICP between non-sequential frames, outlier factors are rejected using a minimum required point cloud overlap, then PCM \cite{pcm}. 
\begin{flalign*}
\mathbf f(\boldsymbol \Theta) = \mathbf  f^{\text{0}}(\boldsymbol \Theta_0) & \prod_i \mathbf f^{\text{O}}_{i}(\boldsymbol \Theta_i) \prod_j \mathbf f^{\text{SSM}}_j(\boldsymbol \Theta_j) \prod_q \mathbf f^{\text{NSSM}}_q(\boldsymbol \Theta_q)
\end{flalign*}
Note that poses (keyframes) are instantiated if the dead reckoning system indicates a distance or rotation larger than a threshold compared to the previous pose. When each robot passes its state to the team members, it incrementally passes its optimized poses as keyframes are instantiated.

\subsection{Distributed Multi-robot SLAM}
Now that each robot has a system to estimate its state, we extend it to include the rest of the team's states in a distributed multi-robot SLAM solution. We use the existing pose graph to integrate inter-robot loop closures relating team members’ trajectories. Note that each robot is responsible for its SLAM solution while passing the required information to build a multi-robot state estimate that includes each robot. 

\begin{figure}[t]
\centering
{\includegraphics[width = .75\linewidth]{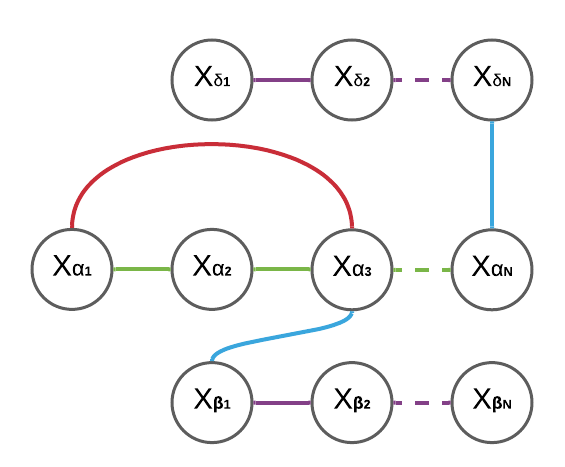}}
\caption{\textbf{SLAM Factor Graph.} Robot poses $\mathbf{x}$ for three robots, $\alpha$, $\beta$, $\delta$ are considered with several factors: sequential scan matching factors (SSM) in green, non-sequential scan matching factors (NSSM, intra-robot loop closures) shown in red, inter-robot (IR) loop closures in blue and partner robot (PR) factors in purple. 
}
\vspace{-6mm}
\label{fig:factor_graph}
\end{figure}

Each robot will incrementally share its pose estimates via the datalink as keyframes are added. We add any detected inter-robot factors, $\mathbf  f^{\text{IR}}$, as in Section V-F. Next, we will add the entire trajectory for each robot in the team if that robot has at least one $\mathbf  f^{\text{IR}}$. The trajectory will be added as a series of sequential factors between robot poses, denoted $\mathbf  f^{\text{PR}}$, \textit{partner robot} factors. This completes the factor graph as shown in Fig. \ref{fig:factor_graph}. 

\subsection{Inter-robot Loop Closure Search}
We formulate the search for inter-robot loop closure candidates as a retrieval problem. Given a candidate and a database, we wish to search the database for possible matches. To do so, we build a scene descriptor using a range-based histogram similar to \cite{thrun-book} for each scene. These histograms encode a scene according to the number of measurements at each range bin, making it rotationally invariant. Note that range bins are discrete, the size and range of which are provided a priori. Using the point cloud, each histogram bin 
is the number of points in its respective range bin. 

Each robot encodes and communicates its scene descriptors to the rest of the team incrementally as the keyframes are instantiated. Note that when this message is shared, it also includes the pose estimate of the keyframe in the sender robot’s reference frame. Additionally, each robot maintains a k-d tree of its scene descriptors. As a robot receives scene descriptors from team members, it compares them to its own using a nearest neighbor search, with a designated maximum neighbor distance. Once nearest neighbors are identified, registration is attempted. 

\subsection{Registration}
When performing registration between two scenes, we estimate a 3DOF rigid body transformation
\begin{align}
     T = \begin{pmatrix} x_{\scriptscriptstyle\Delta}\ 
    y_{\scriptscriptstyle\Delta}\end{pmatrix}^\top
\end{align}
\begin{align}
     R = \begin{pmatrix} \cos{\theta_{\scriptscriptstyle\Delta}}\ -\sin{\theta_{\scriptscriptstyle\Delta}} \\  \sin{\theta_{\scriptscriptstyle\Delta}} \ \;\;\;\;\cos{\theta_{\scriptscriptstyle\Delta}}
    \end{pmatrix},
\end{align}
in order to derive inter-robot factors, $\mathbf  f^{\text{IR}}$. In the registration problem we minimize the squared Euclidean distance between two sets of points $A$ and $B$, that have been associated according to nearest neighbor. 
\begin{align}
     E(R, T) = \Sigma ||a_i - Rb_i + T||^2.
     \label{eq:icp}
\end{align}
Note we do not have any notion of an initial guess between robot reference frames, and therefore have no notion of initial guess when solving the registration problem. It is for this reason we utilize Go-ICP \cite{go-icp}. Go-ICP finds a globally optimal registration result, showing good performance under partial overlap and poor initial guess conditions. For Go-ICP initial alignment, we apply Euclidean mean subtraction to the point clouds and set $ R$ and $T$ as identity and zero, respectively. 
Once Go-ICP completes, we call standard ICP to refine the transformation using the Go-ICP result as an initial guess. 

\subsection{Outlier Rejection}
Once registration is complete, the result must be evaluated as legitimate or erroneous. In our implementation, we consider outlier rejection before and after registration. Before a registration call is made, we check that the point cloud has a minimum number of points, to avoid attempting registration with a cloud containing insufficient information to provide a reasonable transformation estimate. Next, we consider the ratio between the point clouds; if the point ratio is too large or small, the scenes are not likely to be of the same content and therefore discarded. Next, we cast sonar points onto a coarse grid to enable rapid scene-to-scene comparison, which we term the \textit{scene image}. When comparing loop closure candidates, the two relevant scene images are compared, and if the sum of absolute differences between them is high, the scenes are considered different and do not warrant registration. Registration is attempted if the point clouds have the minimum number of points, comply with our ratio requirements, and have a small-enough sum of absolute differences between scene images. 

Once registration is completed, we evaluate the overlap between the two clouds using the resultant transform estimate. Overlap is computed by evaluating the percentage of points with the nearest neighbor inside 0.5 meters. If the overlap is sufficient, we pass the inter-robot loop closure to PCM \cite{pcm} for geometric verification. Once a loop closure is validated by PCM, it is added to the pose graph and communicated to the rest of the robots in the team. 

\subsection{Communication Strategy}
As noted in Section II-C, communication between robots is bandwidth limited. For this reason, we formulate our communications strategy to minimize the transmission of large data structures by first exchanging small ones. After the robot has completed its pose graph optimization at each step, it shares the newest scene histogram, as in Section IV-E, with the associated pose. When another member of the multi-robot system receives this message, the histogram is used to query a k-d tree. If any of the nearest neighbors comply with the maximum tree distance, then that point cloud is requested. Upon receipt of this cloud, registration is attempted as per Sections IV-F and IV-G. This model aims to eliminate the exchange of raw sensor data at each time step. Raw sensor data is only exchanged if scenes are close in feature space, potentially resulting in an inter-robot loop closure. 

Note that we incrementally share pose estimates of robot keyframes as keyframes are instantiated. AUV state estimation, however, is prone to high SLAM drift and subsequent drift correction when loop closures are detected. These corrections mean that robots may need to share updated pose estimates with their team; otherwise, they will be operating with deprecated knowledge of other robots in the system. We handle this by setting a change threshold for a robot's state; if any pose changes significantly, that pose is re-sent to the team. The team then uses this new information to implement the $\mathbf  f^{\text{PR}}$.

\begin{table*}[t]
\centering
\begin{tabular}{ccccccccccc}
\toprule
& \multicolumn{4}{c}{Entire Trajectory}&\multicolumn{4}{c}{Poses With Inter-robot Loop Closures}\\
\midrule
& \multicolumn{2}{c}{Euclidean Dist. (m)}                                & \multicolumn{2}{c}{Theta (deg.)}                                              & \multicolumn{2}{c}{Euclidean Dist. (m)}                                & \multicolumn{2}{c}{Theta (deg.)}                                              & \multicolumn{2}{c}{Network Utilization (bits/second)}                                              
  \\
Case & MAE & RMSE & MAE & RMSE & MAE & RMSE & MAE & RMSE & Average & Min/Max\\
\midrule
SM 1 & 2.83  & 2.89  & 5.12  & 5.12 & 2.2 & 2.2 & 5.12 & 5.12&-&-\\
SM 2  & 2.36 & 2.39 & \bf3.47 & \bf3.47 & 2.06 & 2.06 & 3.47 & 3.47&-&-\\
SM 3  & \bf1.92 & \bf1.98 & 3.5 & 3.51 & \bf1.5 & \bf1.97 & \bf3.27 & \bf3.28 & 337.62 & 217.05 / 468.05\\
SM 4  & 2.63 & 2.68 & 4.74 & 4.74 & 2.03 & 2.04 & 4.75 & 4.75 & \bf161.07 & 134.93 / \bf208.55\\
SM 5  & 7.25 & 7.58 & 20.27 & 26.76 & 6.84 & 6.88 & 14.47 & 15.83 & 213.65 & \textbf{124.26} / 486.54\\

\midrule
USMMA 1 & 2.28 & 2.66 & 3.24 & 3.89  & 1.22 & 1.44 & 1.78 & 2.38&-&- \\
USMMA 2 & 2.09 & 2.41 & 2.21 & 2.63 & 1.13 & 1.39 & 1.38 & 1.78&-&-  \\
USMMA 3 & \bf1.17 & \bf1.29 & \bf1.93 & \bf2.27 & \bf0.66 & \bf0.73 & \bf1.27 & \bf1.47& 3176.61 & 2556.64 / 5579.54\\
USMMA 4 & 1.44 & 1.62 & 2.62 & 3.07 & 0.82 & 0.97 & 1.71 & 2.09 & \bf1244.84 & \bf1050.19 / 1284.24\\
USMMA 5 & 3.09 & 3.59 & 17.76 & 22.66  & 3.26 & 3.31 & 11.51 & 13.33 & 2126.56 & 2079.58 / 2173.16\\
\toprule
\end{tabular}
\caption{\textbf{Real world multi-robot SLAM results.} This table summarizes inter-robot error and network utilization for our five test cases. We analyze inter-robot error in two ways; the entire robot trajectory and only poses with inter-robot loop closures. We report the average network utilization and the min/max of the 10 trials for each test case. Note SM denotes SUNY Maritime.} \vspace{-6mm}
\label{table:slam_results}
\end{table*}

\vspace{-4mm}

\section{Experiments} 
\subsection{Hardware Overview}
In this work, we utilize our customized BlueROV2-Heavy as shown in Fig. \ref{fig:leading_rov}. This vehicle features depth and attitude control implemented with an onboard PixHawk. The sensor payload includes a Rowe SeaPilot DVL, VectorNav VN-100 MEMS IMU, Bar30 pressure sensor, and a twin sonar arrangement. The vertical sonar is a Blueprint Subsea Oculus M1200d, and the horizontal sonar is an Oculus M750d. We use the horizontal sonar as the SLAM perceptual input; with a max range of 30 meters, the sensor has a 20-degree vertical aperture and a 130-degree horizontal field of view. In order to manage sensor data and our SLAM system, we utilize the robot operating system (ROS). 
The computer used to manage all instances of SLAM in our multi-robot system is equipped with an Intel i9-9880H 2.30GHz CPU.

In order to perform multi-robot experiments, we record several datasets and concurrently play them back on a single computer. We use $|N|$ instances of our system during playback to create a multi-robot system. Please note that we play back \textit{real data} (sonar, DVL, IMU) and simulate acoustic communication between robots. To simulate communications between robots, we use the existing ROS message passing system to pass the relevant messages between instances of our system. 

\vspace{-1mm}

\subsection{Metrics}
In order to consider different parameterizations of our system, we use several metrics. Note that there is no ground truth in our data, nor is there enough publicly available sonar data with ground truth information to perform a study of this type. However, since the goal of our system is to provide awareness of the team relative to a single robot, we formulate the following. Each dataset has a stable single robot SLAM solution; this solution is transformed into the frame of each team member. The location of each team member in a common reference frame is determined using manual alignment. We then compare the single-agent SLAM outcome, transformed into each team member's reference frame, to the estimate each robot has built of its team members using our system. We denote this as inter-robot error, characterized with mean absolute error (MAE) and root mean squared error (RMSE). 

Our second metric is network utilization, measured by monitoring the communication channel between robots and measuring total traffic. Note that we exclude network overhead and only report the cost of message contents, as network overhead may be hardware-specific. Network utilization is reported as a time series plot and considering the total usage divided by the mission duration. Note that we perform ten trials for each test case, and results for inter-robot error and network utilization are aggregated across all ten trials. 

\subsection{Multi-robot SLAM Ablation Study}
To validate our system, we consider two environments: SUNY Maritime College in The Bronx, New York, shown in Fig. \ref{fig:leading_sat}, and the United States Merchant Marine Academy (USMMA) in Kings Point, New York. At SUNY Maritime, we use a system of two robots crossing paths from opposite sides of the environment, with a single opportunity for inter-robot loop closure. At USMMA, we consider a system of three robots traveling in similar directions, with several opportunities for inter-robot loop closure. At SUNY Maritime, the robots start on opposite sides of the environment, while at USMMA all robots start close together. 

\begin{figure*}[t]
\centering
\subfloat[\textcolor{Red}{Robot $\alpha$} \label{fig:raw_sonar}]{\includegraphics[width=0.28\linewidth]{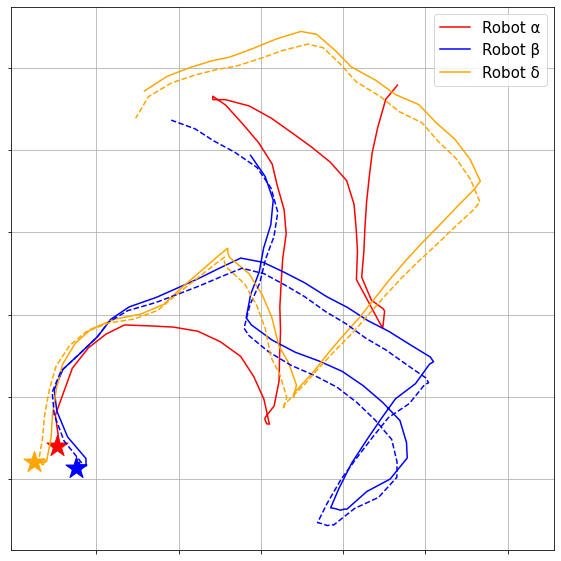}}\ \;
\subfloat[\textcolor{blue}{Robot $\beta$}
 ]{\includegraphics[width=0.28\linewidth]{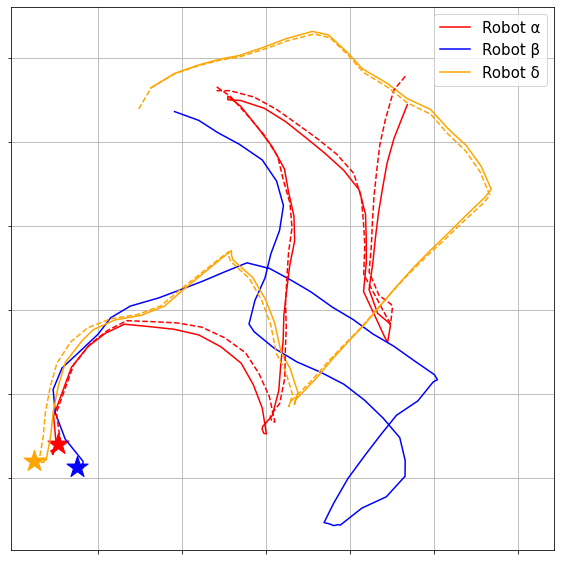}}\ \;
\subfloat[\textcolor{orange}{Robot $\delta$ }\label{fig:pred} ]{\includegraphics[width=0.28\linewidth]{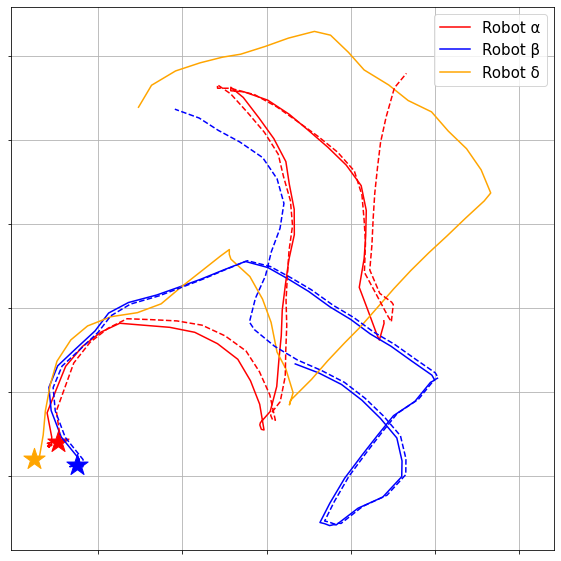}}
\caption{\textbf{Example trajectories from USMMA.} This dataset includes three robots, $\alpha$ (red), $\beta$ (blue) and $\delta$ (orange). Solid lines show trajectory estimates with dotted lines showing ground truth used to compute Table \ref{table:slam_results}. Note that there is only robot relative ground truth, so there is no true (dotted) line for a robot's own trajectory. We also note that estimated trajectories (solid) may be shorter than true (dotted) due to the varying shutdown times of robots. Stars show starting locations for robots. We avoid using $\gamma$ since it refers to sonar intensity. }
\vspace{-4mm}
\label{fig:usmma_trajectories}
\end{figure*}

To test the algorithmic components of our multi-robot SLAM architecture, we present an ablation study in which we progressively add components of our system until the system is complete. Specifically, we vary the contents of our outlier rejection system and specific components of the communication strategy. We note that there is no difference in system parameterization between datasets. For each dataset, we test each of the following cases: \vspace{-1mm}
\begin{itemize}
    \item \textbf{Case 1:} min. no. pts., point ratio, and overlap conditions
    \item \textbf{Case 2:} min. no. pts., point ratio, overlap, and scene image conditions
    \item \textbf{Case 3:} min. no. pts., point ratio, overlap, scene image conditions, and PCM
    \item \textbf{Case 4:} min. no. pts., point ratio, overlap, scene image conditions, PCM, and point cloud compression (\textbf{The complete proposed system})
    \item \textbf{Case 5:} min. no. pts., point ratio, overlap, scene image conditions and PCM without re-sending updated pose information
\end{itemize} \vspace{-1mm}
Our results are summarized in Table \ref{table:slam_results}, and qualitative examples are shown in Fig. \ref{fig:usmma_trajectories}. We also provide playback of our experiments (for case 4) in our video attachment. We require a minimum number of 75 points, a maximum point ratio of 2.0, a scene image max. of 0.8, minimum overlap of 55\% and compression voxel size of 0.25 meters. When computing SLAM metrics, we consider Euclidean distance in meters and yaw (theta) in units of degrees. We break our SLAM metrics into two categories: (1) the entire trajectory and (2) only the poses with inter-robot loop closures. We also report network utilization in units of bits/second for each case. Note that cases 1 and 2 do not have network utilization reported. Their communication configuration does not differ from case 3, nor does their difference in outlier rejection change any form of communication logic. 

\begin{figure}[t]
\centering
{\includegraphics[width = .75\linewidth]{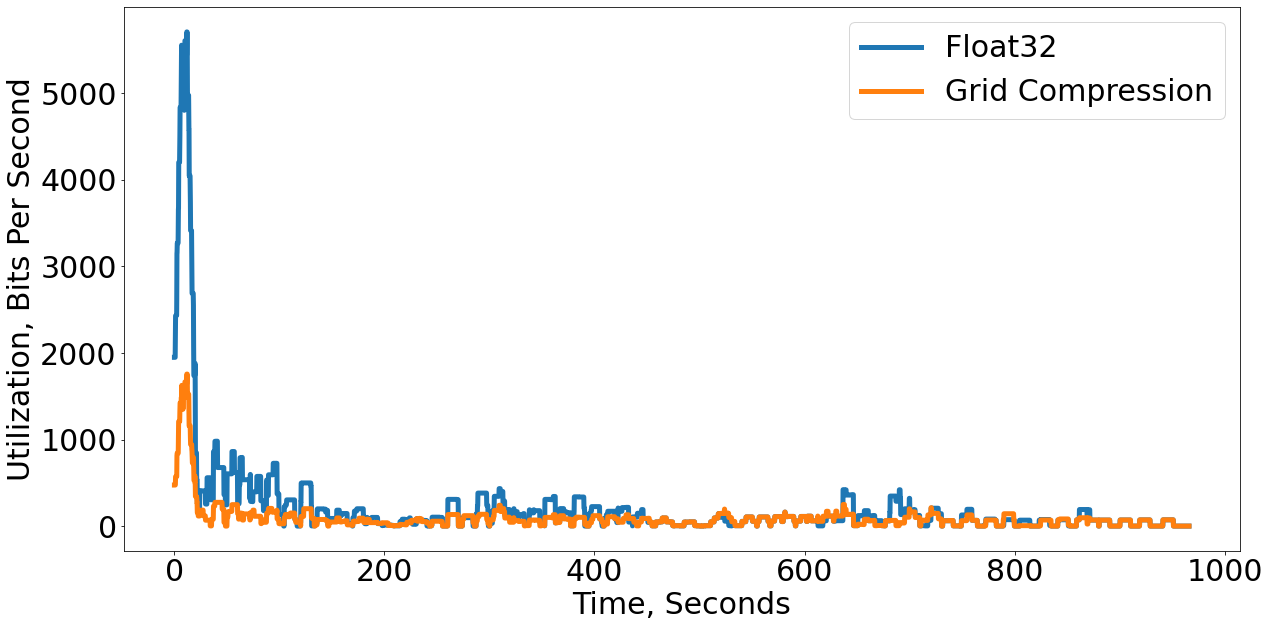}} \\
{\includegraphics[width = .72\linewidth]{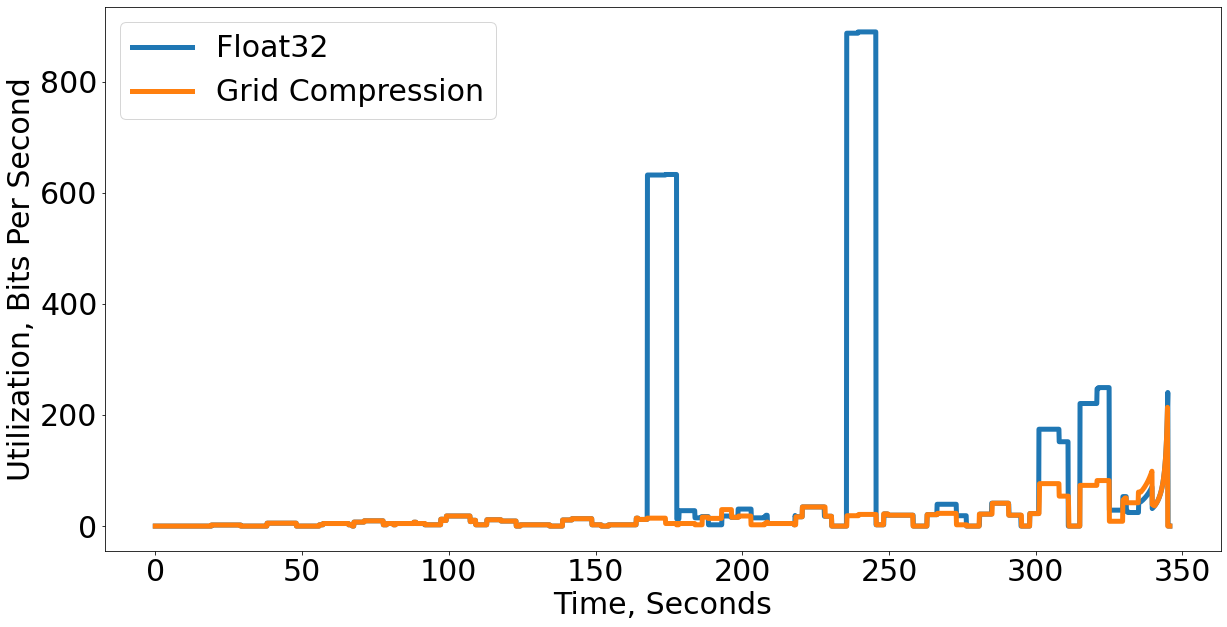}}
\caption{\textbf{Network utilization.} A comparison of network utilization with and without the voxel compression method described in Section IV-B. The y-axis shows network utilization using a sliding window (window of 100) average. The x-axis shows time. A representative run is shown from the three-robot USMMA datset at top, and from the two-robot SUNY Maritime dataset at bottom. 
}
\vspace{-6mm}
\label{fig:network_plot}
\end{figure}

When considering the results of our multi-robot SLAM ablation study, there are several important takeaways. Firstly there is added value in using a scene image as a method of outlier rejection (case 2) at both USMMA and SUNY Maritime in terms of error. Secondly, the addition of PCM (case 3) yields additional value in terms of inter-robot error. Next, when considering the use of point cloud compression (case 4), there is a slight increase in inter-robot SLAM error in exchange for cutting network utilization in half compared to uncompressed clouds (case 3). Time series plots showing network utilization are given in Fig. \ref{fig:network_plot}. We note the reduction in network usage when using the compression method of Sec. IV-B, as well as the spike at the beginning of the USMMA mission, which is due to heavy point cloud exchange. Lastly, when removing the sending of significant updates in a robot's state estimate to the rest of the team (case 5), a greatly deteriorated inter-robot SLAM result is observed with some savings in network utilization relative to case 3. 

We note that our system runs faster than keyframes are added; an aggregate summary of runtimes from USMMA and SUNY Maritime is shown in Table \ref{table:run_time}. We note that our registration time in Table \ref{table:run_time} \textit{includes} our outlier rejection system (without PCM, which is captured separately in the table). It is important to note that network utilization with three agents is reasonable when considering the limitations of commercial off-the-shelf hardware options. We want to underscore the significance of this result. This work is the first instance of multi-robot SLAM using real imaging sonar data to utilize indirect encounters, to the best of our knowledge. Not only does our system result in inter-robot loop closures, but these loop closures are also leveraged to build estimates of team members in the frame of each robot in a distributed manner. Further, it does so with an efficient exchange of information, as shown by the network utilization results. Lastly we note the success rate, i.e., how often inter-robot loop closures are successfully detected. For cases 1-4 in both data sets and USMMA case 5 we note a 100\% success rate. For SUNY Maritime case 5, we note a 70\% success rate, but given this is a data point meant to show the value of an otherwise included feature, we consider it no further. 

\vspace{-1.5mm}

\subsection{Analysis of Perception Message Overhead}
\vspace{-1.0mm}

Some readers may note that using a computer vision paradigm is common when attempting to register a pair of keyframes with no initial guess. This general paradigm extracts salient point features, associates them, and then computes a transform with RANSAC or similar. We note the use of KAZE \cite{kaze} and AKAZE \cite{akaze} features on sonar imagery \cite{Westman-2018,Wang-2019} and the use of ORB \cite{orb} in RADAR imagery. However, in the multi-robot case, that would require broadcasting feature points \textit{and descriptors}. We provide a summary of some relevant feature extraction tools for sonar imagery in Table \ref{table:data_counts}. Here we consider the data-overhead per feature and the number of features per sonar image in our datasets. We use the OpenCV implementations of KAZE, AKAZE, and ORB to perform this comparison. 

\begin{table}[t]
\centering
\begin{tabular}{ccc}
\toprule
Algorithm & Mean (ms) & Standard Deviation (ms)\\
\midrule
PCM & 0.32 & 0.33\\
Kd-tree Search & 0.88 & 4.49\\
Registration (w/ rejection) & 258.88 & 265.43\\
\midrule
\end{tabular}
\caption{\textbf{Runtime.} PCM is used for loop closure verification, k-d tree search is for scene comparison, and registration refers to the Go-ICP based registration method in Section IV-F, including the outlier rejection computations.} 
\vspace{-2mm}
\label{table:run_time}
\end{table}

\begin{table}[t]
\centering
\begin{tabular}{cccc}
\toprule
Method & Mean Count & Std. Dev. & Mean KBits \\
\midrule
Scene Descriptor & 1 & 1 & .128 \\
Point cloud - float32 & 138.69 & 142.94 & 9.14\\
Point cloud - compressed &  138.69 & 142.94 & \bf 2.28\\
KAZE & 48.68 & 56.73 & 116.20\\
AKAZE & 37.05 & 41.14 & 20.07\\
ORB & 168.62 & 322.43 & 82.54\\
\midrule
\end{tabular}
\caption{\textbf{Perception message overhead.} Mean count and std. dev. refer to the average number of points extracted with the given method. Mean Kbits are computed by taking the overhead of a single point and multiplying it by the mean number of points.} 
\vspace{-6mm}
\label{table:data_counts}
\end{table}

As shown in Table \ref{table:data_counts}, our scene descriptor is a data-efficient means of summarizing content. Moreover, even passing simple 32-bit floating-point-based point clouds is significantly less costly than KAZE, AKAZE, and ORB. Lastly, our compression method, while simple, shows a considerable reduction in data requirements, and as we have shown in Section V-C and Table \ref{table:slam_results}, only results in a minor increase in inter-robot error.

\vspace{-2mm}

\subsection{Multi-agent Mapping}
While the purpose of our system is to provide each robot with knowledge of the states of the other agents in the system, it can also be used to derive merged maps. Consider a system of robots tasked with mapping an area in a collaborative modality, a core task for a multi-robot system. 
This section showcases an example of a map that was merged offline, using the multi-robot SLAM results from Section V-D. We specifically consider the SUNY Maritime dataset, which has two robots. Using the multi-robot state estimate, we can transform the sonar point clouds in the frame of each robot, shown in Fig. \ref{fig:leading_1}. We note the coverage increase with the use of a multi-robot system and the drift in our SLAM system. This drift is primarily due to the high drift rate of our low-cost MEMS IMU. We also note that the inter-robot drift is comparable to that of a single robot sweeping the entire workspace shown in Fig. \ref{fig:leading_1}.

\vspace{-2mm}

\section{Conclusions} 
\vspace{-1mm}
In this work, we have proposed a system to find and integrate inter-robot loop closures in a distributed, graph-based pose SLAM architecture. We have demonstrated the real-time viability of our system, the accuracy of accepted inter-robot constraints, and the efficacy of our communications system. When considering the potential shortcomings of our system, while we take steps to prevent perceptual aliasing from corrupting the SLAM solution, this effect can overwork the communications link. If scene descriptors are similar, we exchange point clouds, meaning that perceptual aliasing can result in the over-exchange of information that our outlier rejection system will cull. Future work will consider geometric verification in scene descriptors to ensure only useful perceptual data is exchanged. However, this is the first example of imaging sonar-based underwater multi-robot SLAM, to the best of our knowledge. There is much potential for future work, including achieving a wide variety of cooperative tasks, and multi-robot active SLAM. Furthermore, our system enables multi-agent autonomous operations in settings where GPS is unavailable or clock synchronization is impractical, which are relevant constraints in many real-world applications.


\vspace{-3.5mm}

{}

\end{document}